\documentclass[letterpaper, 10 pt, conference]{ieeeconf}  

\IEEEoverridecommandlockouts               

\usepackage{xcolor}
\usepackage{placeins}
\usepackage{tcolorbox}
\usepackage{listings}
\usepackage{booktabs}

\usepackage{algorithm}
\usepackage{algpseudocode}
\usepackage{amsmath}
\title{\LARGE \bf MindArm: Mechanized Intelligent Non-Invasive Neuro-Driven Prosthetic Arm System}

\author{Maha Nawaz$^{*}$ and Abdul Basit$^{*}$ and Muhammad Shafique
\thanks{Maha Nawaz, Abdul Basit, and Muhammad Shafque are with the eBRAIN Lab, Division of Engineering, New York University (NYU) Abu Dhabi, United Arab Emirates, {\tt\footnotesize \{mzn2386, abdul.basit, muhammad.shafique\}@nyu.edu}. $^{*}$Authors have equal contribution}
\thanks{This work was supported by Rachmad V. W. Putra, New York University (NYU) Abu Dhabi, United Arab Emirates, {\tt\footnotesize rachmad.putra@nyu.edu}}
}

\usepackage{graphicx} 
\usepackage{caption}
\usepackage{caption}
\usepackage{threeparttable} 
\renewcommand{\thetable}{\Roman{table}}

\begin{document}

\maketitle
\thispagestyle{empty}
\pagestyle{empty}

\begin{abstract}

Currently, individuals with arm mobility impairments (referred to as ``\textit{patients}'') face limited technological solutions due to two key challenges: (1) non-invasive prosthetic devices are often prohibitively expensive and costly to maintain, and (2) invasive solutions require high-risk, costly brain surgery, which can pose a health risk. Therefore, current technological solutions are not accessible for all patients with different financial backgrounds.
Toward this, we propose a low-cost technological solution called \textit{MindArm}, an affordable, non-invasive neuro-driven prosthetic arm system. \textit{MindArm} employs a deep neural network (DNN) to translate brain signals, captured by low-cost surface electroencephalogram (EEG) electrodes, into prosthetic arm movements. Utilizing an Open Brain Computer Interface and UDP networking for signal processing, the system seamlessly controls arm motion. In the compute module, we run a trained DNN model to interpret filtered micro-voltage brain signals, and then translate them into a prosthetic arm action via serial communication seamlessly. Experimental results from a fully functional prototype show high accuracy across three actions, with 91\% for idle/stationary, 85\% for handshake, and 84\% for cup pickup. The system costs approximately \$500-550, including \$400 for the EEG headset and \$100-150 for motors, 3D printing, and assembly, offering an affordable alternative for mind-controlled prosthetic devices.

%
\end{abstract}

\begin{keywords}
EEG, OpenBCI, non-invasive, prosthetic arm, deep learning, low cost.
\end{keywords}

\section{Introduction}
\begin{figure}[t]
    \centering
     \includegraphics[width=1\linewidth]{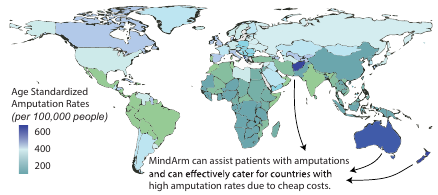}
    \caption{Global Incidence of Traumatic Amputations. Based on the work by Yuan et al. \cite{10.3389/fpubh.2023.1258853} published in Front. Public Health, this map illustrates the age-standardized amputation rates across 204 countries and territories. Areas with higher ASIRs could benefit significantly from the MindArm prosthetic solution, which offers a cost-effective and accessible alternative to assist individuals who have experienced limb loss, thus addressing the global need for affordable prosthetic care.}
    \label{fig:motivation}    
\end{figure}

Globally, millions of people live with disabilities, including limb amputations, which severely limit their ability to perform everyday tasks. In recent years, approximately 5.4 million people live with paralysis in the United States alone \cite{armour} and 57.7 million people live with limb amputations globally \cite{mcdonald}; see the global distribution of age-standardized amputation rates in Fig.~\ref{fig:motivation}. Among them, there are people with disability or difficulty to move their arms, which we refer to as ``\textit{patients}'' in this paper for brevity. 
These patients have limited access to effective technological solutions for managing their physiological limitations, primarily because many current innovations are not affordable for individuals from diverse financial backgrounds.

 Existing prosthetic solutions, particularly mind-controlled devices, are prohibitively expensive, often exceeding \$100k \cite{Kwek2016IsAP, Beyrouthy2016EEGMC}. Invasive solutions also exist but carry additional costs and risks associated with surgery \cite{Huinink2016LearningTU,collinger}. To address the need for a cost-effective, non-invasive solution, we propose \textit{MindArm}, a system that utilizes widely available surface EEG electrodes to control a prosthetic arm via a DNN engine.


Hence, there is a significant need for \textit{an alternative low-cost solution that help the patients to perform diverse activities}.

\textit{\textbf{Targeted Research Problem:}
How can we develop a low-cost solution that can help the patients to move their arms for performing desired actions?}
An efficient solution to this problem will help the patients from different financial backgrounds to access an alternate low-cost solution for addressing their physiological limitations (i.e., unable to move hands) and performing diverse activities, thereby improving their quality of life. 

\subsection{State-of-the-Art Solutions and Their Limitations}

Various state-of-the-art prosthetic solutions exist, but they are typically expensive and not accessible to all patients. Below is a comparison between \textit{MindArm} and leading commercial prosthetic arms.

\begin{table}[h!]
\centering
\footnotesize
\caption{Comparison of MindArm with Other Prosthetic Solutions}
\begin{tabular}{@{}lllll@{}}
\toprule
\textbf{Prosthetic Arm}       & \textbf{Cost (USD)}           & \textbf{Features/ Maintenance}                                                        \\ \midrule
\textit{\textbf{MindArm (ours)}}              & 500 - 550                      & EEG-based, 3D-printed, Low                       \\
\textit{DEKA Arm System}      & 100K - 200K                     & EMG-based, High                      \\
\textit{Ottobock} & 60K - 120K                    & Advanced bionic arm, Medium               \\
\textit{Bebionic Arm}         & 25K - 75K                     & Multi-articulating hand, Medium                    \\
\textit{i-Limb Quantum}       & 40K - 60K                       & Advanced multi-grip, Medium                    \\ \bottomrule
\end{tabular}
\end{table}

To address the physiological limitations of the patients, the existing solutions can be loosely classified into two categories: (1) \textit{non-invasive solutions}, such as surface mounted eeg electrode devices; (2) and \textit{invasive solutions}, such as implantable brain-computer interface (BCI).

\textbf{Non-invasive Solutions:} 
These solutions do not require invasive devices inside the patients' body to obtain high quality information signals. 
For instance, prosthetic-based solutions like 
``DEKA bionic arm''~\cite{bloomer2020comparison} use electromyography (EMG) signals to capture the muscle signals and then translate them into the desired action. 
Another non-invasive solutions (e.g., gesture control armband) focus on capturing and interpreting muscle signals without physical actuator attached~\cite{Musfequr}, thereby making them lightweight.
In such EMG-based solutions, when the patients contracts their muscles, the electrodes detect the muscle signals, then send them to the compute module for further processing~\cite{Yildiz}~\cite{Visconti}. 
However, EMG signals may not be present in amputated patients, or cases where physical conditions are severely compromised (e.g., patients with severe disability, such as paraplegia and tetraplegia/quadriplegia). 
%
Furthermore, the other non-invasive solutions like mind-controlled prosthetic devices are typically very costly (i.e., over \$100K) and hence expensive to maintain~\cite{Kwek2016IsAP}\cite{Beyrouthy2016EEGMC}. 
It is partially due to the effective yet expensive materials such as titanium alloys~\cite{sarraf2021state}.

\textbf{Invasive Solutions:} 
To improve the quality of information signals, some solutions employ invasive devices to access the signal sources.   
Consequently, this device needs to be physically implanted inside the patients' body through invasive surgery. 
However, such an approach is a high risk procedure to perform and typically very expensive. 
In addition, most of such technologies are not wireless, thereby making them difficult to maintain. 
For instance, to acquire electroencephalogram (EEG) waves, ones need to put electrodes in the patients' head (skull), and attach the plug and connecting cables to the compute module. 
Furthermore, such invasive-based solutions are not commercially available, and it is estimated that when it becomes available, it may cost hundreds of thousands of dollars~\cite{sandra}. 
Hence, not all patients with different financial backgrounds can afford that. 

In summary, the existing technological solutions are still very expensive as well as difficult to maintain. 
Moreover, in the invasive solution cases, there is a high risk to the patients' body, which may cause other negative side effects. 

Given the benefits and weaknesses of the state-of-the-art, we identify that the potential solution is to consider a non-invasive solution with low-cost technology, while ensuring high accuracy of signal processing that can correctly interpret the input signals into actions. 
To fulfill such requirements, we opt to develop \textit{a non-invasive EEG-based prosthetic arm}. 



\subsection{Scientific Challenges}
 
Our non-invasive EEG-based prosthetic arm solution bears potentials to address the existing weaknesses in the state-of-the-art, but it also imposes several challenges, as discussed in the following.
\begin{itemize}
    \item To reduce the design cost, one of the main challenge is to design a system with an effective signal processing pipeline, that can be implemented using a low-cost off-the-shelf devices and modules. 
    %
    \item There is environmental noise which pollutes the EEG signals. 
    Such artefacts need to be removed, and thereby requiring an effective denoising process.  
    \item It requires an effective algorithm to learn and extract information from EEG signals. Once the system learns the EEG features, it should be able to correlate these features to the corresponding prosthetic arm action.
\end{itemize}

\begin{figure}[t]
    \centering
    \includegraphics[width=1\linewidth]{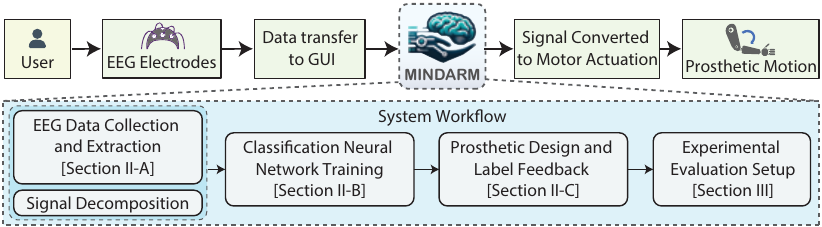}
    \caption{Our MindArm Workflow: EEG electrodes capture the brains' micro-voltage signals, which are then collected by the brain-computer interface (BCI) system with GUI. The data undergoes decomposition, cleaning, and temporary cloud storage before classification by the neural network. After training, the system predicts intended actions in real-time, and these actions are communicated to the prosthetic limb, enabling it to move accordingly.}
    \label{fig:introduction}    
\end{figure}

\subsection{Our Novel Contributions}
To address the targeted problem and scientific challenges, we propose \textit{\textbf{MindArm}}, a mechanized intelligent non-invasive neuro-driven prosthetic arm system; see an overview in Fig.~\ref{fig:introduction}. 
It employs a deep neural network (DNN) engine to extract information from EEG signals for identifying the given instruction, then translate it into a prosthetic arm action. 
To achieve this, our MindArm system makes the following novel contributions.
\begin{itemize}
    \item \textbf{Removing Noise in the EEG Signals (Section~\ref{sec:DataExtraction}):}
    We reduce noise in the EEG signals by extracting band power from each channel with respective frequencies such that residual noise is filtered, as well as employing metal insulation to reduce eddy currents and domestic alternate current noise. 
    %
    \item \textbf{Learning EEG Signal Features using DNN Training (Section~\ref{sec:NeuralNetTrain}):}
    We employ DNN training to effectively learn EEG signal features that are obtained from a low-cost off-the-shelf Ganglion board. 
    We employ a window buffer size to overcome the shortcomings of a small number of EEG channels on the Ganglion board. 
    %
    %
    \item \textbf{Low-Cost Prosthetic Arm Design (Section~\ref{sec:ProstheticLabel}):}
    We design the prosthetic arm structure in Fusion360, and build it using 3D printer and Prusa MK3 with a combination of ABS\footnote{ABS: Acrylonitrile Butadiene Styrene}, PETG\footnote{PETG: Poly-Ethylene Terephthalate Glyco}, and PLA\footnote{PLA: Poly-Lactic Acid} filaments to provide a low weight yet strong structure.
    Lastly, the design of the prosthetic is modular, allowing all parts can be easily replaced if required, thereby reducing the maintenance cost.
    %
\end{itemize}

\begin{figure*}[ht]
    \centering
    \includegraphics[width=1\linewidth]{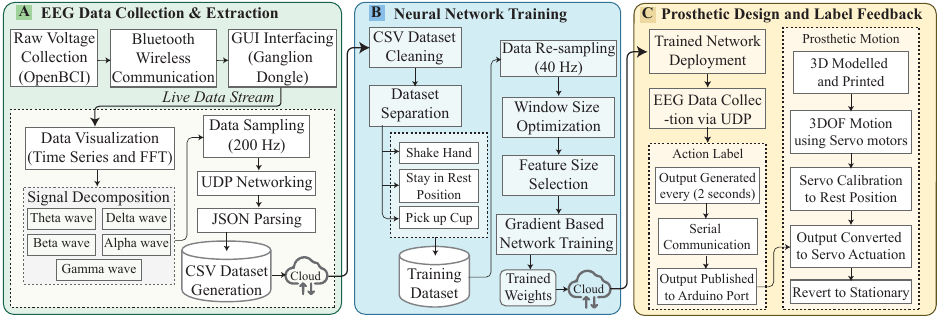}
    \caption{MindArm Methodology:
    (a) Data Collection: EEG voltages are transmitted via Bluetooth to the GUI, which processes the data through a Fast Fourier Transform (FFT). The decomposed data are then saved to the cloud for further processing.
    (b) Neural Network Training: The dataset undergoes cleaning and is categorized into three groups. To accommodate UDP communication latency, the sampling rate is adjusted to 40 Hz. A window size of 80 samples, including a feature set of 20, is selected for training the network, which is subsequently saved to the cloud.
    (c) Deployment: The trained network interfaces with the prosthetic, designed with servo motors to enable 4 Degrees of Freedom (4DoF) movement. Network outputs are converted into servo actuations to perform the intended tasks.}
    \label{fig:methodology}    
\end{figure*}

\textbf{Key Results:}
To evaluate our MindArm, we build a complete setup encompassing EEG acquisition module, compute module (i.e., DNN engine), and actuator module (e.g., servo motor and prosthetic arm) allowing for for 3 degrees of freedom, whose total cost is $\sim$\$550. 



Our fully functional prototype of the MindArm system demonstrates promising success rates in performing three predefined actions, showcasing the efficacy of MindArm as an affordable solution for a mind-controlled prosthetic arm.

\section{Methodology}
In this section we will describe the MindArm system in detail, along with the dataset generation, refinement and training workflow. 
We also elaborate the prosthetic design and label feedback selection along with system integration; see an overview in Fig.~\ref{fig:methodology}.

\subsection{EEG Data Collection \& Extraction}
\label{sec:DataExtraction}

The off-the-shelf state-of-the-art devices on the market for collecting EEG signals include the OpenBCI complete Ultracortex \cite{OPENBCI}, Emotiv EPOC X, and Flex kit. 
However, in addition to the price of the Emotiv kits \cite{Emotiv}, they also require preparation of either saline soaked felt or gel coating further decreasing practical effectiveness compared to the dry electrodes utilized with the Ganglion brain computer interface \cite{OPENBCI}. 
\textit{Therefore, to maintain a low price for the prosthetic and maintain practicality of dry electrodes, the Ganglion board is utilized}. 

The OpenBCI Ganglion board features 4 EEG channels and 2 references, facilitating the use of both dry comb electrodes and flat electrodes with a snap connection interface coated in silver-silver chloride \cite{OPENBCI_board}. 
Flat electrodes are positioned at Fp1 and Fp2 locations in the \textit{nasion}, and comb electrodes at T3 and T4, with references placed at A1 and A2. 
The placement at Fp1 and Fp2 aims to enhance alpha wave detection. 
Although alpha waves are more prominent in the \textit{occipital lobe}, placement near the \textit{inion} (back of the head) often encounters greater noise. Consequently, while O1 and O2 locations yield stronger alpha and theta wave signals, achieving stable amplitudes across different frequencies using Fast Fourier Transform (FFT) analysis proves more challenging. 
Evidently, a scientific challenge arises when attempting to interpret data from only 4 EEG channels compared to numerous EEG channels in non-invasive state of the art devices. Fig. 4 \& 5 shows user wearing the headset.

\begin{figure}[ht]
  \begin{minipage}{0.5\linewidth}
    \includegraphics[width=\linewidth]{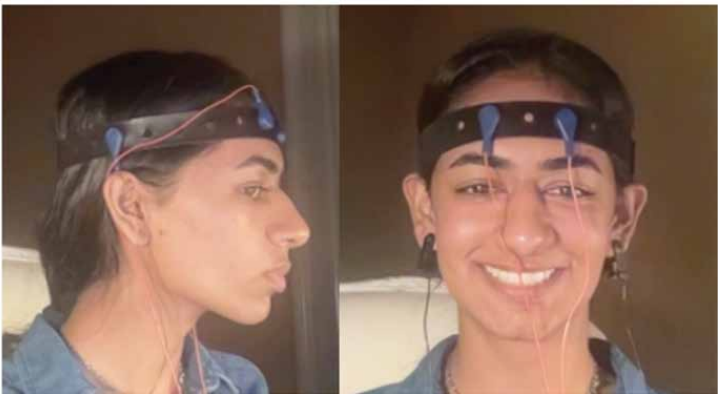} 
    \caption{The electrodes are attached to an adjustable band that has different slots for peripheral EEG readings.}
  \end{minipage}%
  \hfill 
  \begin{minipage}{0.45\linewidth}
  \centering
    \includegraphics[width=0.55\linewidth]{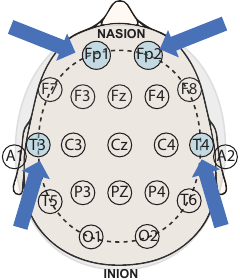} 
    \caption{10-20 system for electrode placement, highlighting the region FP1, FP2, T3, and T4 being used for the measurements.}
  \end{minipage}
\end{figure}

\begin{table}[ht]
\centering
\caption{Comparison of EEG Devices}
\label{tab:eeg_devices}
\begin{tabular}{lccc}
\hline
Device Name & Price & Channel/Sensors & Requirements \\
\hline
\textbf{Ganglion BCI (ours)} & \$400 & 4 channels & None \\
Emotiv EPOC X & \$800 & 14 sensors & Saline/Gel \\
Emotiv Flex Kit & \$1700 & 32 sensors & Saline/Gel \\
openBCI Ultracortex & \$2400 & 16 channels & None \\
\hline
\end{tabular}
\end{table}


\begin{figure}[ht]
    \centering
    \includegraphics[width=0.9\linewidth]{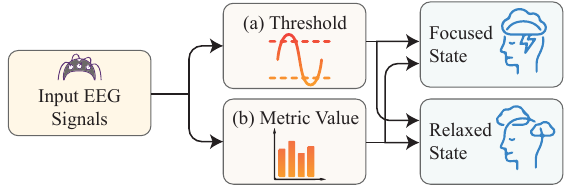}
    \caption{ EEG Signal Processing Overview. (a) Population-based neural network thresholds indicate average band power values, enhancing general algorithm accuracy at the expense of individual specificity. (b) Metric values, derived from these thresholds, facilitate relaxation state detection through FFT-based amplitude criteria across gamma, beta, alpha, theta, and delta brainwave bands.}
    \label{fig:threshold}    
\end{figure}

\begin{figure}[ht]
    \centering
    \includegraphics[width=1\linewidth]{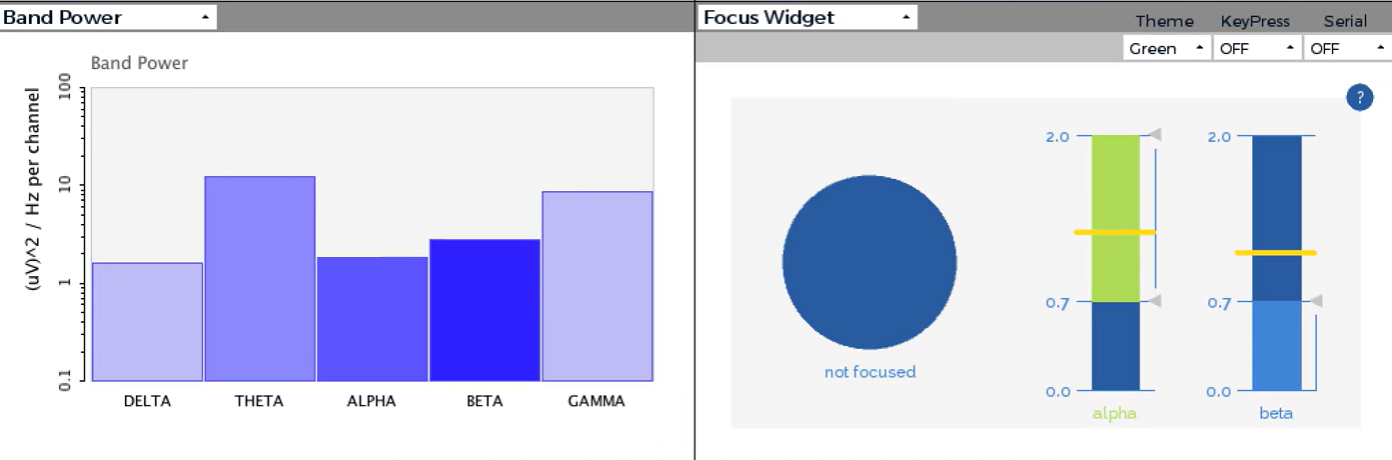}
    \caption{The left panel displays EEG band power across Delta, Theta, Alpha, Beta, and Gamma frequencies. The right panel (Focus Widget) presents the subject's cognitive state. The large circle indicates the overall focus status, where "not focused" is shown. Adjacent bar graphs depict the threshold comparison for Alpha and Beta waves, indicating the subject’s relaxation (Alpha) and concentration (Beta) levels relative to preset thresholds (yellow line), helping classify the subject's mental state.}
    \label{fig:threshold}    
\end{figure}


In the initial phase of the study, rather than directly decomposing the EEG signal into specific frequency bands, as shown in Fig. 6, a more rudimentary threshold-based system \cite{chen2021dual} was employed to ascertain the subject’s state of relaxation or concentration, as previously used in EEG signal processing for cognitive state detection. The determination of the subject being relaxed or focused was then mapped to corresponding actions to be executed by the prosthetic.

The results indicate promising trends; however, the current method presents a notable limitation: the states of concentration and relaxation are not distinctly categorized by Graphical User Interface (GUI) shown in Fig. 7.
Consequently, the algorithm fails to differentiate between `relaxed' and `concentrated' states, only recognizing `relaxed and stationary' as well as `concentrated and stationary' states. This issue stems from an observable overlap in the metric thresholds for relaxation and concentration within the GUI. Therefore, it is imperative to develop an alternative approach that enables accurate prediction across all three desired states: \textit{handshaking}, \textit{cup grasping}, and \textit{remaining stationary}. 
Notably, a parallel in amplitude across various frequencies is observed when comparing the thought of handshaking with the relaxed state, and the thought of cup grasping with the concentrated state as shown in Fig. 8.

\begin{figure}[ht]
  \begin{minipage}{0.5\linewidth}
    \includegraphics[width=\linewidth]{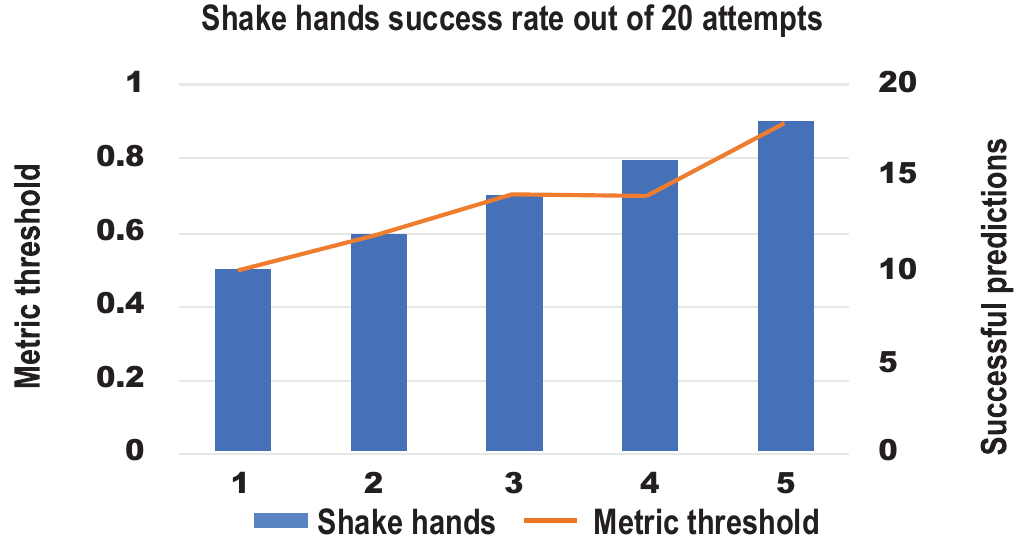} 
    \end{minipage}%
  \begin{minipage}{0.5\linewidth}
    \includegraphics[width=\linewidth]{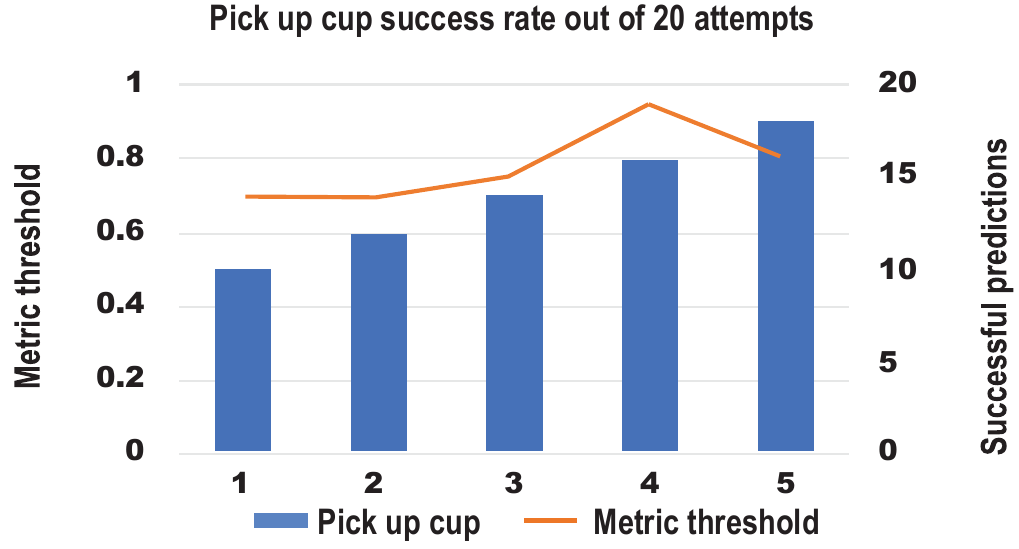} 
  \end{minipage}
\caption{Correlation between metric threshold and prediction success rates. Panel (a) Illustrates a positive relationship where an increased metric threshold is associated with a higher number of successful handshaking predictions out of 20 attempts. (b) shows a generally positive trend in the number of successful predictions for cup grasping as the metric threshold rises; however, a notable decrease in successful predictions is observed at a threshold of 0.9, under-performing the success rate at the threshold of 0.8.}
\end{figure}

To enhance the accuracy of state prediction, the proposed methodology involves adopting a strategy of associating relaxation with the intent to shake hands and concentration with the intent to pick up a cup. See Fig. 10 \& 11.


To validate the efficacy of this approach, a novel signal processing technique is employed, which decomposes EEG signals into five distinct brain wave categories: \textit{delta}, \textit{theta}, \textit{alpha}, \textit{beta}, and \textit{gamma}, each within their characteristic frequency bands: delta (1–4 Hz), theta (4–8 Hz), alpha (8–13 Hz, with sub-bands alpha-1 8–10 Hz and alpha-2 11–13 Hz), beta (13–30 Hz), and gamma (above 30 Hz) as illustrated in Fig. 9.
This spectral decomposition allows for more precise extraction of features from each channel, thereby reducing noise transmission through the User Datagram Protocol (UDP) network and enhancing algorithmic accuracy. 
The choice of UDP is driven by its expedited data transfer capabilities, absence of connection establishment procedures, and greater efficiency via lower bandwidth usage and overhead, thus compensating for potential latency and ensuring a robust set of features for the neural network classifier, contributing to a low-maintenance system  \cite{Roshdy2019}.

\begin{figure}[ht]
  \begin{minipage}{0.55\linewidth}
    \includegraphics[width=\linewidth]{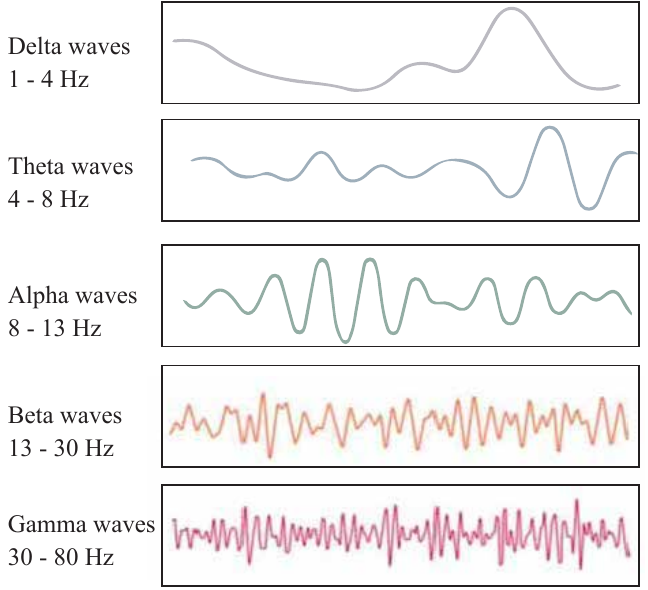} 
  \end{minipage}%
  \hfill 
  \begin{minipage}{0.38\linewidth}
    \caption{Characteristic Brain Wave Patterns. Gamma waves exhibit peaks during problem-solving and deep concentration. Beta waves are pronounced with active mental engagement. Alpha waves are elevated during periods of relaxation and reflection. Theta waves increase with drowsiness, while delta waves are predominant in deep sleep.}
    \end{minipage}
\end{figure}

\begin{figure}[ht]
    \centering
    \includegraphics[width=1\linewidth]{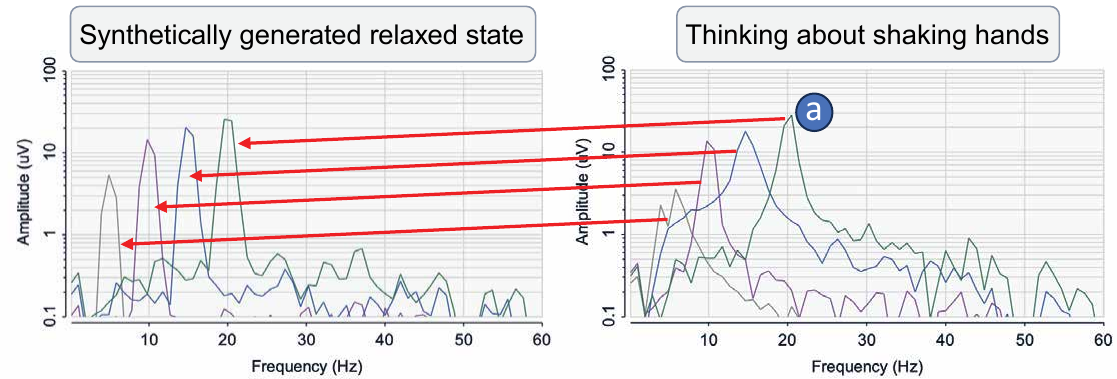}
    \caption{Similarity in Amplitude Spikes Across Frequencies. The
    Fast Fourier Transform (FFT) graphs display coinciding amplitude
    peaks at corresponding frequencies (a), indicating analogous band
    power characteristics among the channels.}
    \label{fig:result_5}    
\end{figure}

\begin{figure}[ht]
    \centering
    \includegraphics[width=1\linewidth]{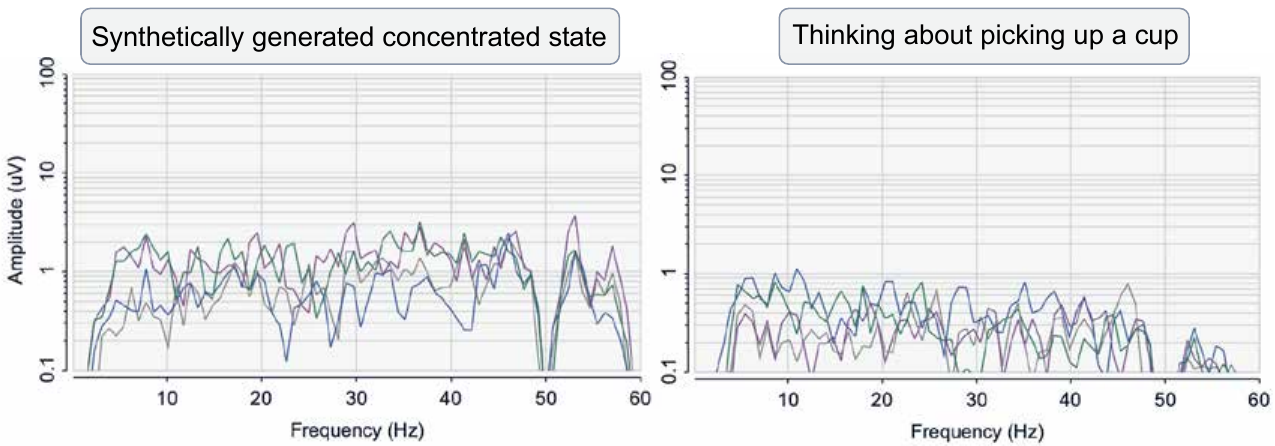}
    \caption{The graphs illustrate the frequency amplitudes across the spectrum. Notably, the amplitude pattern associated with a concentrated state closely matches that observed during the action of picking up a cup, suggesting that this action is predominantly executed in a state of concentration.}
    \label{fig:result_6}    
\end{figure}

Moreover, the system capitalizes on the ganglion board's capability to process data at a sampling rate of 200Hz by transmitting only the most significant bit from each EEG sample \cite{OpenBCI_format}. This transmission approach results in slightly lower sample resolution, which is considered a negligible trade-off. Consequently, the neural network receives a substantial, numerically labeled dataset every second for each classification category. This enhances the learning rate of the prosthetic actions, offering a cost-effective and faster training methodology when compared to other BCI devices within the same price bracket of \$400, such as the ganglion.


\begin{figure}[ht]
    \centering
    \includegraphics[width=0.9\linewidth]{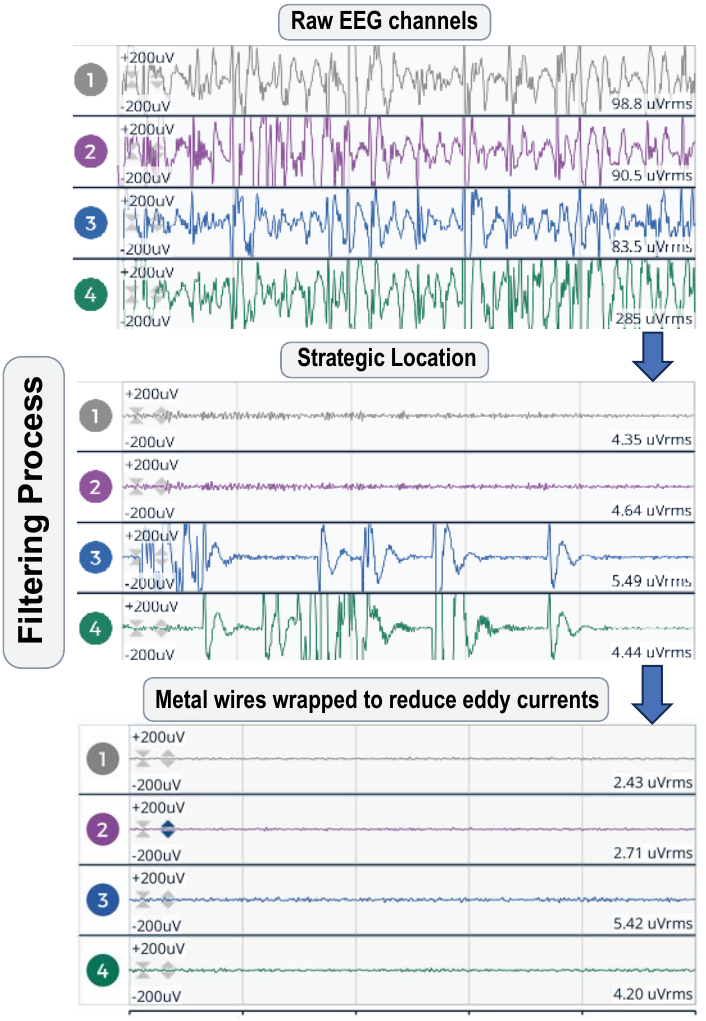}
    \caption{EEG Signal Cleaning Process: Illustrates the application of novel techniques for the removal of major artifacts, resulting in the accurate representation of amplitudes at specified frequencies.}
    \label{fig:result_7}
\end{figure}

A significant impediment encountered with non-invasive mind-controlled prosthetics is environmental electrical noise interference at the user's location. To mitigate this, data was sampled across various settings with differing levels of domestic alternating current (AC). Areas with minimal AC interference exhibited reduced noise in the EEG channels. Despite this, the reduction was not adequate to yield clean channel features suitable for neural network input. Moreover, this approach does not align with the practical need for portability, as the prosthetic should function optimally in diverse locations irrespective of AC noise levels. The results of this filtering process are shown in Fig. 12.



With artifacts minimized, it is essential to transmit the band power data to the Python IDE in real-time. This transmission is facilitated through UDP networking, enabling data transfer via a designated port and socket. The maximum buffer size is set to 1024 bytes, which sufficiently accommodates the data payload. For each sample frame per second, the data is converted from binary to decimal format, flattened, and then written to a designated file, where all the data within that file is categorized under the same class.

The flattening process is represented as:
\texttt{flattened\_data = [i] + [item for sublist in data\_list for item in sublist]}

Here, \texttt{`i'} represents the input number, facilitating continuous tracking of the dataset size.



\subsection{Neural Network Training}
\label{sec:NeuralNetTrain}


Currently, the algorithm processes three datasets, each corresponding to a distinct action executed by the prosthetic:
\begin{itemize}
    \item Shaking hands with an individual.
    \item Remaining stationary or in a resting position.
    \item Picking up a cup.
\end{itemize}

\begin{figure}[ht]
    \centering
    \includegraphics[width=0.8\linewidth]{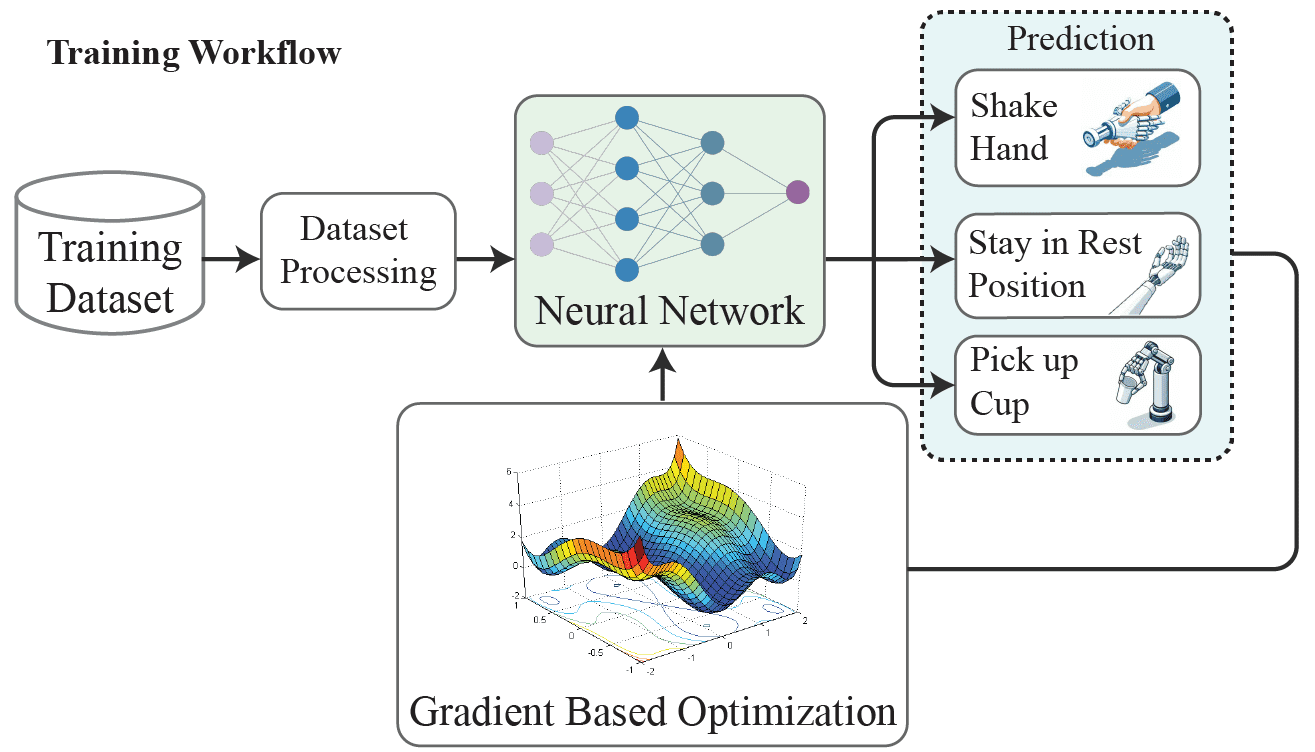}
    \caption{Cleaned data from the files are fed into the neural network, which then undergoes its training process. During this process, the labels of the inputs are compared with the neural network's predicted results, facilitating gradient descent optimization via the loss function. This optimization adjusts the network's parameters to improve performance.}
    \label{fig:training_workflow}
\end{figure}


During the data collection phase, the user is instructed to concentrate on one of three predetermined actions for a fixed duration. Concurrently, the acquired data undergoes cleaning procedures before being stored in a CSV file, with each entry tagged with a numerical label corresponding to the envisioned action. For instance, data associated with the act of shaking hands is recorded in `shakehands.csv' and labeled as `0'. Although the intended sampling rate is 200Hz, practical limitations due to latency in data processing result in an effective rate of approximately 50Hz. This discrepancy necessitates an extended duration for the training phase. The resultant datasets comprise between 10,000 and 20,000 sampled pieces, employed for training the neural network. The data structure includes 20 columns, reflecting the aggregation of five normalized brainwave metrics across four distinct EEG channels.






\begin{algorithm}
\caption{Data preprocessing and dynamic segmentation into windows with label assignment}
\begin{algorithmic}[1]
\small
\State \textbf{Input:} Datasets $df_{pickUpCup}$, $df_{shakeHands}$, $df_{stayIdle}$ with corresponding labels 0, 1, 2
\State \textbf{Output:} Windowed features $X_{windowed}$, One-hot encoded labels $y_{windowed}$

\Procedure{SegmentDataset}
{$X, y, win\_size, num\_chan$}
    \State $max\_overlap \gets max win\_size/2$
    \State $min\_overlap \gets min(20, win\_size/4)$
    
    \State Initialize $segmented\_X \gets []$, $segmented\_y \gets []$
    \State $start\_idx \gets 0$
    \While{$start\_idx + win\_size \leq X.shape[0]$}
        \State $overlap \gets$ random integer between $min\_overlap$ and $max\_overlap$
        \State $step\_size \gets win\_size - overlap$
        \State $end\_idx \gets start\_idx + win\_size$
        \State $segment \gets X[start\_idx:end\_idx, :]$
        \State $segment \gets segment.reshape(win\_size, num\_chan)$
        \State Append $segment$ to $segmented\_X$
        \State Append $y[start\_idx]$ to $segmented\_y$
        \State $start\_idx \gets start\_idx + step\_size$
    \EndWhile
    \State \textbf{return} $segmented\_X$, $segmented\_y$
\EndProcedure

\State Combine and preprocess datasets 
\State Assign labels and standardize features
\State Dynamically segment features into windows with random overlap and assign labels
\State Combine windowed data from all actions
\State Split data into training and test sets
\State Initialize $EEGDataset$ with features and labels
\State Create $DataLoaders$ for training and test sets

\end{algorithmic}
\end{algorithm}

In the exploration of optimal neural network architectures for our dataset, a diverse range of models was assessed.
These included simple Feedforward Neural Networks (FFNNs), Recurrent Neural Networks (RNNs), Long Short-Term Memory networks (LSTMs), CNN-LSTM hybrids, and Transformer-based networks. Among these, the Transformer-based network outperformed other models achieving an impressive validation accuracy of 97.1\%. This superior performance can be attributed to the Transformer's ability to process sequential data in parallel and its efficient handling of long-range dependencies, which are critical for understanding complex patterns within the EEG dataset. However, the trade-off for this high level of accuracy involves increased computational resources and training time, compared to simpler models like FFNNs or RNNs. The decision to employ a Transformer-based model thus reflects a strategic balance between seeking optimal performance and managing the computational costs associated with more sophisticated architectures. The transformer architecture is illustrated in Fig. 14. The training accuracy of these networks are depicted in Fig. 15 showcasing the comparative training performance across epochs for the various network architectures. The transformer model showcases the highest validation accuracy. Fig. 16 shows window size optimization.

\begin{figure}[ht]
  \begin{minipage}{0.4\linewidth}
    \includegraphics[width=\linewidth]{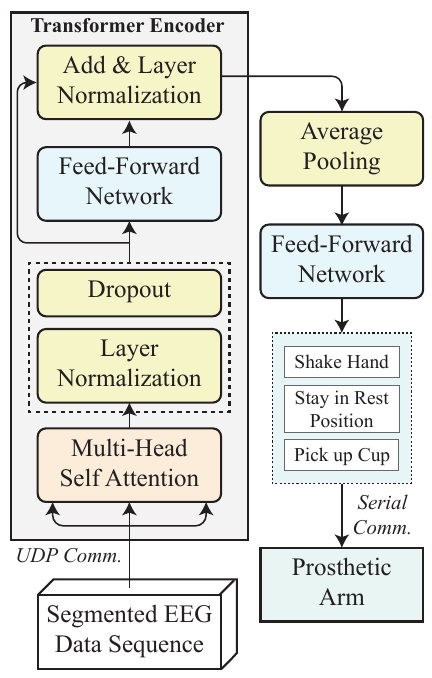} 
  \end{minipage}%
  \hfill 
  \begin{minipage}{0.48\linewidth}
    \caption{ EEG Transformer Architecture. The depicted Transformer-based model processes segmented EEG data sequences, leveraging a multi-head self-attention mechanism and subsequent layer normalization within the Transformer Encoder. Data flows through an average pooling layer before passing through two feed-forward networks, culminating in the classification of the prosthetic arm's actions. These predicted actions are then communicated to the prosthetic arm via serial communication for execution.}
  \end{minipage}
\end{figure}

\begin{figure}[ht]
    \centering
    \includegraphics[width=1\linewidth]{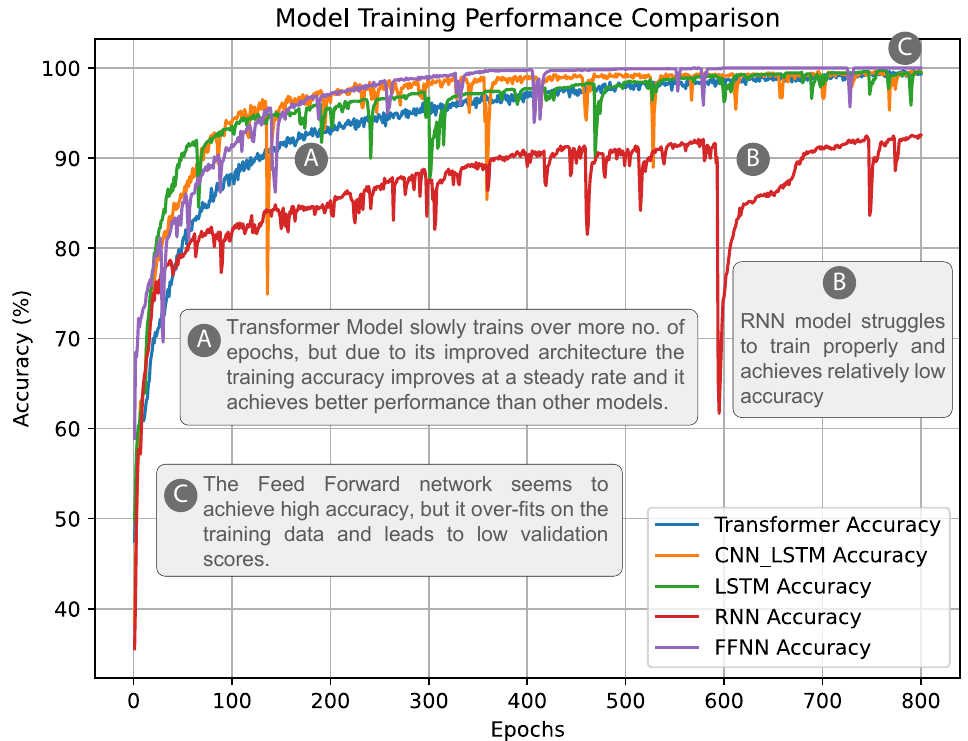}
    \caption{Comparative Training Performance of Neural Network Architectures. This graph presents the training accuracies over numerous epochs for each network model tested. It highlights the Transformer-based network’s superior performance in accuracy relative other networks, underlining its robust capability in learning from the EEG dataset.}
    \label{fig:training}
\end{figure}


Initial training sessions revealed challenges with label volatility, as the model output action labels at a rate of approximately 40 times per second over UDP. This high frequency of label generation led to instances where transient thought patterns inadvertently triggered unintended actions. For example, a brief, unintended contemplation of an action could result in the erroneous activation of the prosthetic arm.

To address this, the model's training and input collection were adapted to include a larger window size, enhancing data stability and output accuracy. It is critical for the input dataset to accurately reflect sustained thought patterns associated with specific actions, which typically last more than a fortieth of a second. In practice, a thought duration of at least 2 seconds is necessary for consistent brain wave intensity.

\begin{figure}[ht]
    \centering
    \includegraphics[width=0.8\linewidth]{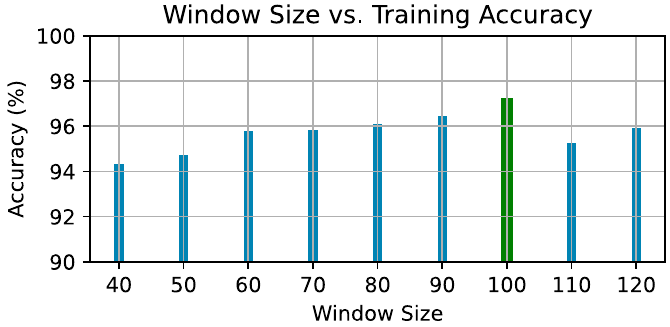}
    \caption{Impact of Window Size on Transformer Network Accuracy. This bar graph illustrates the relationship between various window sizes and the corresponding accuracy of the model. Each bar represents a different window size which is dynamically overlapped. The highest accuracy marked by a green bar underscores the optimal window size that maximizes accuracy, highlighting the importance of window size selection in neural network performance.}
    \label{fig:window}
\end{figure}

Accordingly, an optimized window size of 100 was implemented. This setup accumulates 100 rows of CSV data—each row representing a 20-dimensional vector from the EEG—into a single tensor. This tensor is then reshaped into a 1x2000 matrix (20x100), serving as the input for the neural network. This approach ensures that the input data effectively represents approximately two seconds of EEG data, allowing for more accurate and representative model outputs as live EEG data is streamed as shown in Fig. 17.

 
 
\begin{figure}[ht]
    \centering
    \includegraphics[width=0.8\linewidth]{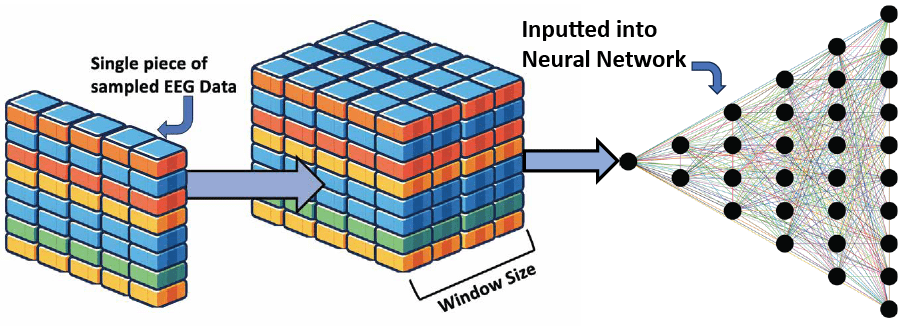}
    \caption{EEG Data Sample Preparation. Each EEG data sample, comprising 20 features, is incorporated into a dynamically sized window, accumulating multiple samples for neural network input. This window size of data will be inputted into the neural network as a single input with each ‘block’ being an individual feature utilized for the neural network training.}
    \label{fig:training_workflow}
\end{figure}


This approach necessitates that the neural network receives a singular, consolidated input for processing. Consequently, the total number of potential inputs from each file was determined, followed by the reshaping of matrices to conform to the neural network's specified input dimensions


\begin{table}[h]
\centering
\caption{Dataset Dimensions Before and After Processing}
\label{tab:dataset_dimensions}
\begin{tabular}{lcc}
\hline
Action & Original Dimension & Windowed Dimension \\
\hline
Shake Hands & $[13697, 20]$ & $[136, 100, 20]$ \\
Stay Stationary & $[19622, 20]$ & $[196, 100, 20]$ \\
Pick Up Cup & $[20837, 20]$ & $[208, 100, 20]$ \\
\hline
\end{tabular}
\end{table}


The refactored algorithm transmits the user's intended action to the Arduino every 2 seconds, improving model robustness by considering more features over a sustained period. Pre-training the neural network and storing it on the cloud ensures faster label output than the input interval, minimizing data loss during wireless EEG streaming.

\subsection{Prosthetic Design and Label Feedback}
\label{sec:ProstheticLabel}


The prosthetic control system transmits the output label to the Arduino via serial communication every 2 seconds. At a standardized baud rate, the Arduino interprets the received number `0, 1, or 2' and initiates the corresponding action. To prevent potential damage to the servos and artificial tendons due to rapid oscillation between positions, the algorithm is designed to pause reading incoming data until the current action is at least a third way through executing.


\begin{figure}[ht]
  \begin{minipage}{0.4\linewidth}
    \includegraphics[width=\linewidth]{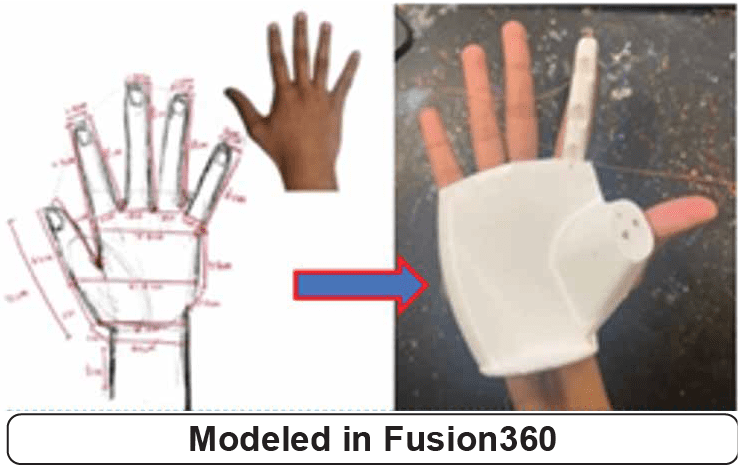} 
  \end{minipage}%
  \hfill 
  \begin{minipage}{0.55\linewidth}
    \caption{The prosthetic was designed and modeled in Fusion360 to the precise dimensions of the author's hand, incorporating 5 tendon tubes for threading braided fishing line, as illustrated in the adjacent figure.}
  \end{minipage}
  \label{fig:hand}
\end{figure}

Initially, the prosthetic's actuation mechanism relied on the contraction and relaxation of tendons, facilitated by a servo horn. However, this design was found impractical due to the persistent friction between the tendons, made of braided fishing line, and the joint pins. Consequently, this friction led to the tendons' degradation over time, resulting in both wear and tear of the prosthetic components and a decline in performance. See Fig. 18 - 21.

 
\begin{figure}[ht]
    \centering
    \includegraphics[width=1\linewidth]{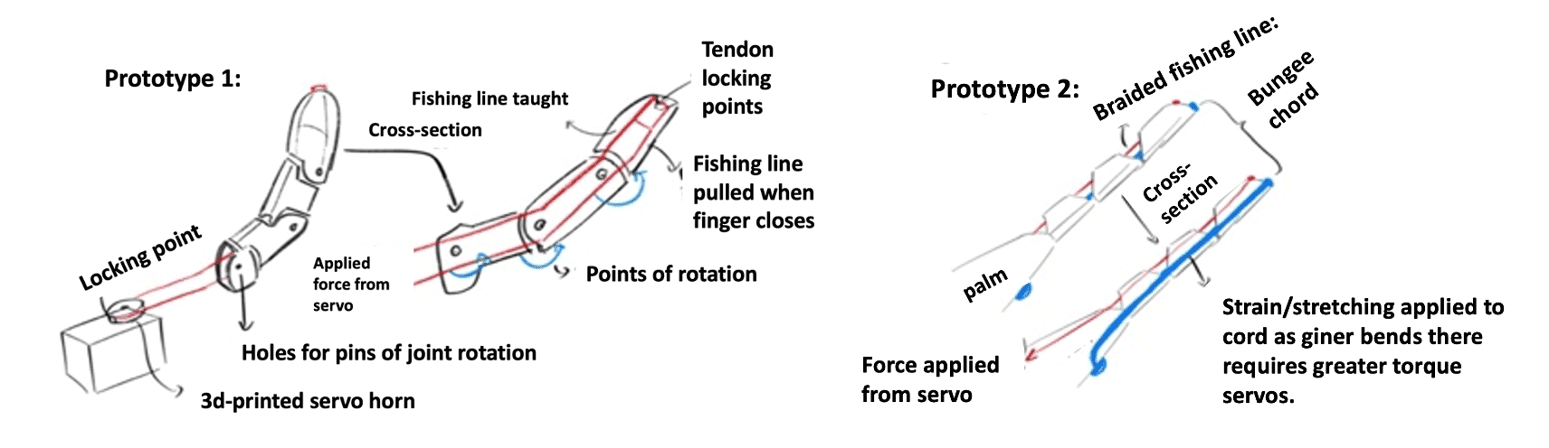}
    \caption{ The second prototype introduces elastic bungee cords to return the finger to its original position, replacing the previous mechanism of coupled flexion and extension via tendons. The use of bungee cords minimizes friction during stretch and contraction, unlike the movement of braided fishing lines, thereby enhancing the prosthetic's durability}
    \label{fig:hand_model}
\end{figure}


20 kg torque servos are utilized at the elbow joint to ensure the load borne by the prosthetic is adequately supported, thereby guaranteeing system durability and practicality. A modular design was developed for each servo compartment, allowing users to easily replace servos without the need for specialized tools.

\begin{figure}[ht]
    \centering
    \includegraphics[width=1\linewidth]{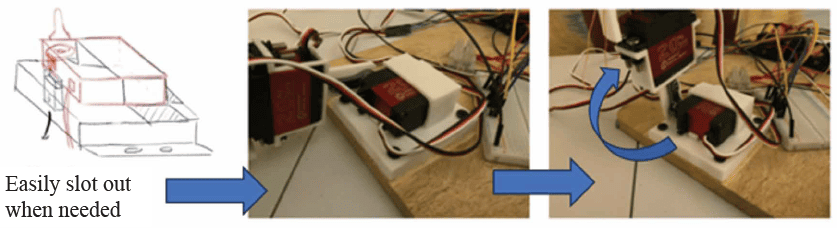}
    \caption{Modular Elbow Servo Joint. This design feature enhances the prosthetic's maintainability, allowing for easier repairs and upgrades.}
    \label{fig:servo}
\end{figure}

In the current market, commercially available realistic prosthetic gloves, exemplified by products from companies like Ottobock, typically start at a price point exceeding 250 USD. In contrast, this study utilized a silicone mold technique to fabricate a comparable realistic prosthetic glove at a material cost of just 15 USD. This substantial reduction in cost represents a significant stride toward democratizing access to prosthetic technology, markedly lowering the financial barrier for potential users.


\begin{figure}[ht]
    \centering
    \includegraphics[width=1\linewidth]{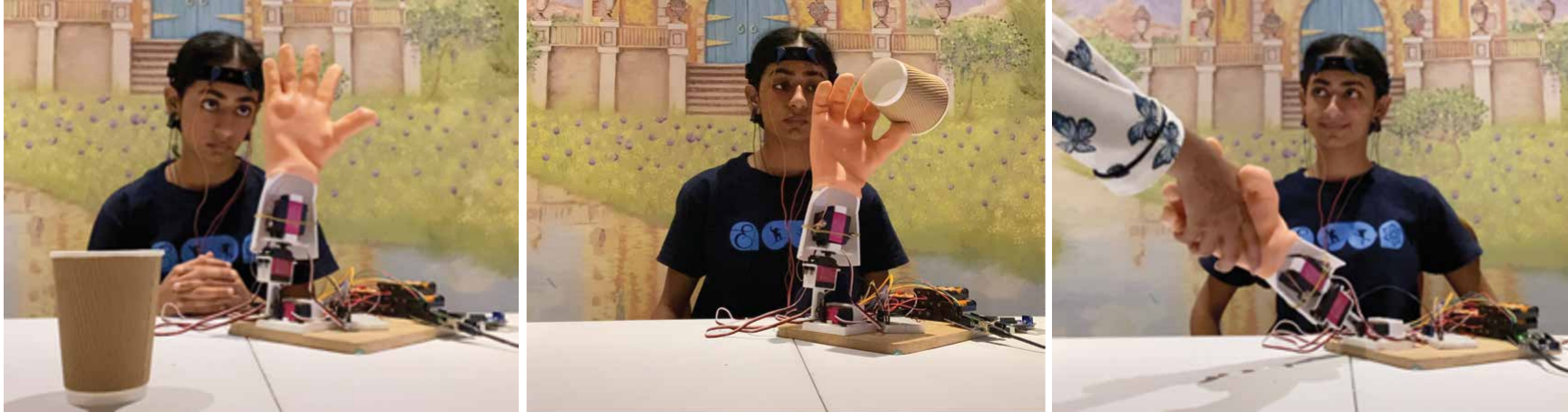}
    \caption{Demonstration of the prosthetic functionality. The left panel depicts the user concentrating on the action of picking up a cup, prompting the prosthetic to mimic the gesture in the center frame. The right panel shows the user contemplating a handshake, which the prosthetic executes accordingly.}
    \label{fig:demo}
\end{figure}


\section{Results and Discussion}
\subsection{Experimental Setup}


The experimental setup includes the Ganglion board with OpenBCI GUI for neural signal acquisition and UDP networking for data communication. Neural network development is done using PyTorch, with commands transmitted via Arduino C++ IDE. Design and prototyping use Tinkercad, Fusion360 for 3D modeling, and Prusa Slicer for 3D printing and circuit refinement.

\subsection{Analysis}

The Transformer network's was analysed by interfacing the model output with an Arduino board using serial communication protocols. The Arduino was programmed to translate the neural network's output into actionable commands for the prosthetic hand. Each predicted action from the network triggered the corresponding movement in the prosthetic hand, showcasing the potential of this system in real-world applications. The deployment of this system demonstrated not only the high accuracy of the Transformer network, as reflected in the classification report and the confusion matrix but also its capability to operate in real time with the physical hardware, offering a seamless transition from prediction to action execution.

\begin{figure}[ht]
  \begin{minipage}{0.75\linewidth}
    \includegraphics[width=\linewidth]{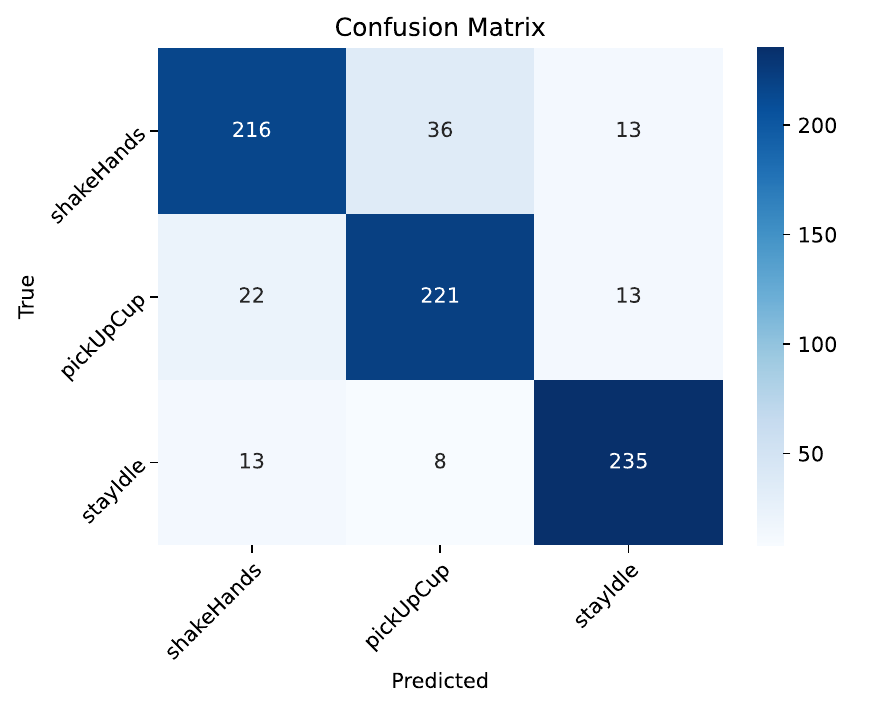} 
  \end{minipage}%
  \hfill 
  \begin{minipage}{0.20\linewidth}
    \includegraphics[width=\linewidth]{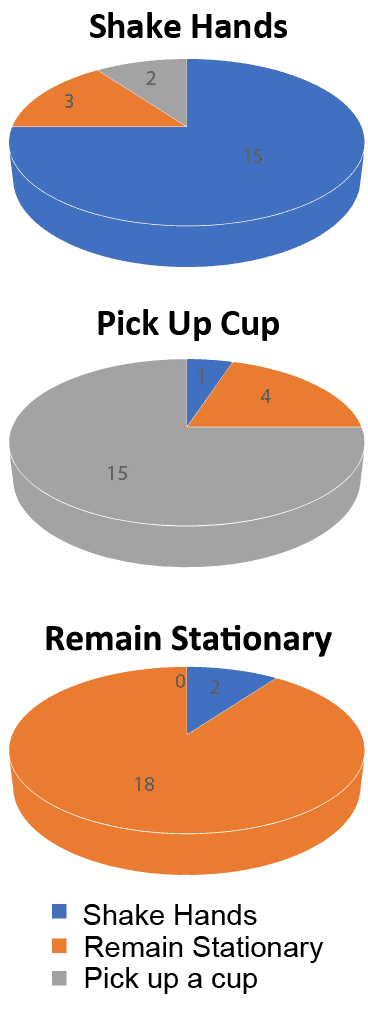} 
  \end{minipage}
  \caption{The confusion matrix shows actual and predicted actions, showing the number of instances for each action for the Test Dataset. The matrix highlights the model's ability to differentiate between the prosthetic's actions with a high degree of accuracy on Test Dataset, The pie chart shows the real world test cases and it showcases good accuracy in these tasks.}
\end{figure}

\begin{table}[H]
\centering
\caption{Classification Report}
\label{tab:classification_report}
\begin{tabular}{lcccc}
\hline
Class & Precision & Recall & F1-Score \\ 
\hline
pickUpCup & 0.86 & 0.82 & 0.84  \\
shakeHands & 0.83 & 0.86 & 0.85  \\
stayStationary & 0.90 & 0.92 & 0.91  \\
\hline
Accuracy & & & 0.86  \\
Macro Avg & 0.86 & 0.87 & 0.86 \\
Weighted Avg & 0.86 & 0.86 & 0.86  \\
\hline
\end{tabular}
\end{table}

\section{Conclusion}

In this paper, we introduced the MindArm methodology, a low-cost, non-invasive, mind-controlled prosthetic arm designed to assist individuals with disabilities in performing everyday tasks. The system translates brain signals into arm motion using EEG technology, with a deep neural network (DNN) interpreting these signals to control prosthetic actions. Experimental test results demonstrate high success rates: 90\% for idle/stationary, 80\% for shaking hands, and 80\% for picking up a cup, validating the effectiveness of the system as an affordable alternative to existing prosthetic solutions shown in Fig. 22. While MindArm is cost-effective and accessible, it faces limitations such as noise in surface EEG signals and a restricted range of actions. The hardware, though affordable, may also have durability issues. Future efforts will focus on refining signal accuracy, expanding action capabilities, and enhancing hardware robustness through improved control mechanisms and signal filtering.

\section*{Acknowledgment}
This work was partially supported by the NYUAD Center for Artificial Intelligence and Robotics (CAIR), funded by Tamkeen under the NYUAD Research Institute Award CG010.



{\small
\bibliographystyle{IEEEtran}
\bibliography{IEEEfull,cite}
}
\addtolength{\textheight}{-12cm}   


\end{document}